%% file: sdm_main.tex
\documentclass[twoside,leqno,twocolumn]{article}

\usepackage[letterpaper]{geometry}

\usepackage{ltexpprt}
\usepackage{hyperref}
\usepackage{soul}
\usepackage{multirow}
\usepackage{lscape}
\usepackage{graphicx}
\usepackage{subfigure}
\usepackage{adjustbox}
\usepackage{enumitem}
\usepackage{tabularx}
\usepackage{caption}
\usepackage{booktabs}
\usepackage{amsfonts}
\usepackage{algorithmic} 
\usepackage{amsthm}
\usepackage{amsmath}
\theoremstyle{definition}
\newtheorem{definition}{Definition}

\newtheorem{problem}{Problem}

\newcommand{\name}{{\texttt{MedAttacker}}}

\begin{document}

\title{MedAttacker: Exploring Black-Box Adversarial Attacks on Risk Prediction Models in Healthcare}
\author{Muchao Ye\thanks{The Pennsylvania State University. (\{muchao, junyu, ting, fenglong\}@psu.edu)}
\and Junyu Luo \footnotemark[1]
\and Guanjie Zheng\thanks{Shanghai Jiao Tong University.  (gjzheng@sjtu.edu.cn)}
\and Cao Xiao\thanks{Amplitude. (danicaxiao@gmail.com)}
\and Ting Wang\footnotemark[1]
\and Fenglong Ma\footnotemark[1]
}

\date{}

\maketitle

% Copyright Statement
% When submitting your final paper to a SIAM proceedings, it is requested that you include
% the appropriate copyright in the footer of the paper.  The copyright added should be
% consistent with the copyright selected on the copyright form submitted with the paper.
% Please note that "20XX" should be changed to the year of the meeting.

% Default Copyright Statement
%\fancyfoot[R]{\scriptsize{Copyright \textcopyright\ 2022 by SIAM\\
%Unauthorized reproduction of this article is prohibited}}

% Depending on which copyright you agree to when you sign the copyright form, the copyright
% can be changed to one of the following after commenting out the default copyright statement
% above.

%\fancyfoot[R]{\scriptsize{Copyright \textcopyright\ 20XX\\
%Copyright for this paper is retained by authors}}

%\fancyfoot[R]{\scriptsize{Copyright \textcopyright\ 20XX\\
%Copyright retained by principal author's organization}}

%\pagenumbering{arabic}
%\setcounter{page}{1}%Leave this line commented out.

\begin{abstract} \small\baselineskip=9pt Deep neural networks (DNNs) have been broadly adopted in health risk prediction to provide healthcare diagnoses and treatments.  To evaluate their robustness, existing research conducts adversarial attacks in the white/gray-box  setting where model parameters are accessible. However, a more realistic black-box adversarial attack is ignored even though most real-world models are trained with private data and released as black-box services on the cloud. To fill this gap, we propose the first black-box adversarial attack method against health risk prediction models named {\name} to investigate their vulnerability.
%~\footnote{The implementation code is at \url{https://drive.google.com/drive/folders/1GZhDfOOsFLEyXGiUSmRv957YpPFtKYMU?usp=sharing}.} 
{\name} addresses the challenges brought by EHR data via two steps: hierarchical position selection which selects the attacked positions in a reinforcement learning (RL) framework and substitute selection which identifies substitute with a score-based principle. Particularly, by considering the temporal context inside EHRs, it initializes its RL position selection policy by using the contribution score of each visit and the saliency score of each code, which can be well integrated with the deterministic substitute selection process decided by the score changes. In experiments, {\name}  consistently achieves the highest average success rate and even outperforms a recent white-box EHR adversarial attack technique in certain cases when attacking three advanced health risk prediction models  in the black-box setting across multiple real-world datasets. In addition, based on the experiment results we include a discussion on defending EHR adversarial attacks.
\end{abstract}

%\noindent \textbf{Key words.} Black-box Adversarial Attacks, Electronic Health Records, Health Risk Prediction.

\input{subfiles/introduction}
\input{subfiles/relatedwork}

\input{subfiles/method}
\input{subfiles/experiments}

\input{subfiles/conclusions}

\bibliographystyle{siam}
\bibliography{sample-base}

\end{document}

%% file: subfiles/introduction.tex
\section{Introduction}

The increasingly accumulated electronic health records (EHR) data have advanced the field of health analytics, especially the \emph{health risk prediction}~\cite{DBLP:conf/cikm/MaYXCZG18, choi2016retain, song2017attend,luo2020hitanet} task, which aims to predict future health status of patients according to their historical EHR data and nowadays is commonly conducted by deep neural networks (DNNs). Although the robustness of DNN-based health risk prediction systems is of great importance due to its relation with medical resources planning and human lives, the research on \textit{adversarial attack} which quantifies the robustness of DNNs by generating adversarial examples to fool the victim models has not received as much attention on EHR data as on image or text data. Specifically, there is a gap between the settings of existing adversarial attack approaches and real-world application scenarios on health risk prediction models.

%due to the restrictions on sharing proprietary information (e.g., the disease model and patient data used in model training). .. However, the connected EHR systems and risk prediction models are vulnerable for \textit{adversarial attacks} for a number of financial reasons. 
%Recent years have witnessed unprecedented progress on applying deep learning techniques in different application domains such as computer vision, natural language processing (NLP), and healthcare. One of the key tasks in the healthcare domain is \textit{risk prediction}, which aims to predict future health status of patients according to their historical electronic health record (EHR) data. 
% Risk prediction is relevant to many questions in clinical medicine, public health, and epidemiology, and the predicted risks of a specific diagnosis or health outcome can be used to support decisions by patients, doctors, health policy makers, and academics.
% Thus, several commercial health companies such as KenSci\footnote{\url{https://www.kensci.com/}}, Lumiata\footnote{\url{https://www.lumiata.com/}}, and IQVIA\footnote{\url{https://www.iqvia.com/}} have their own products or systems to target this task using state-of-the-art deep learning techniques. 

%Since good understanding of the vulnerability of deep risk prediction models can help  proactively prevent the high-risk attacks, researchers start to investigate the EHR adversarial attack. 

To illustrate, existing studies mainly explore the robustness of deep health risk prediction models by \textbf{white/gray-box adversarial attacks}, which assume attackers can access the parameters of health risk prediction models. For example, Sun et al.~\cite{sun2018identify} propose a white-box one to identify susceptible locations in clinical time series data, An et al.~\cite{an2019longitudinal} generate adversarial EHR examples in the white/gray-box setting, and Wang et al.~\cite{wang2020attackability} test a white-box evasion attack on EHRs. However, in the real world health analytics companies %such as KenSci, Lumiata, and IQVIA 
train their models with their private data and release them as \textbf{black-box} services on the cloud. Therefore, the assumptions of white/gray-box settings are often invalid in real-world practice because the parameters of proprietary models of compnaies are inaccessible. Thus, it is desirable to have a \textbf{black-box adversarial attack} method for understanding the robustness of deep health risk prediction models, but it is difficult to design one owing to  the following challenges.

\begin{figure}[t]
    \centering
    \includegraphics[width=0.4\textwidth]{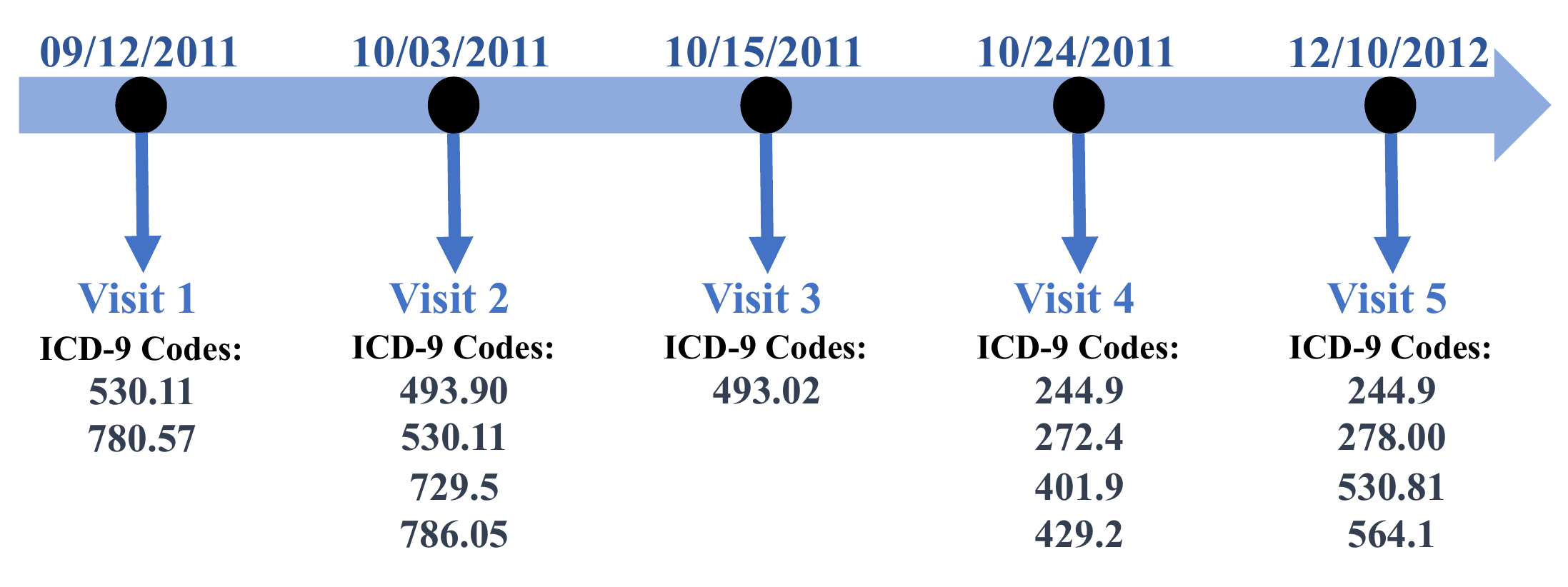} 
    \vspace{-0.15in}
    \caption{Illustration of EHR data. }
    \label{fig:data}
    \vspace{-0.15in}
\end{figure}

%\smallskip
%\noindent 
\textbf{Challenges. \label{Sec: data challenge}}
The challenges stems from the unique structure of EHR data. As shown in Figure~\ref{fig:data}, compared to text data which is also discrete, EHR is different in its \textbf{unordered} diagnosis codes within each visit and the \textbf{incidental} nature of hospital visits. That is, each longitudinal EHR is a sequence of patient visits that show their evolving conditions as well as sporadic incidences, where the diagnosis codes of each visit are unordered. Such uniqueness of EHR raises the challenge of designing adversarial example generation for EHR. Existing black-box techniques attacking discrete data such as text usually  use either score-based methods~\cite{ren2019generating,gao2018black, li2019textbugger}, %which quantify the sensitivity level of each word by removal, 
or reinforcement learning ones~\cite{zang2020learning} to find sensitive words and select replacement words. %which learn an policy for selection and substitution operations. %Despite their good performance in the text domain, 
However, their applicability on generating EHR adversarial example is limited: for one thing, score-based approaches determine replacements directly by the calculated saliency scores, and the generated adversarial examples are likely to be locally optimum; for another, existing reinforcement learning methods~\cite{zang2020learning} have not taken the temporal context into consideration when taking actions. Thus, this technical challenge raises the question of \textit{how to alleviate these limitations to design a black-box adversarial attack approach against health risk prediction models that takes the properties of EHRs into consideration.}

%they either only focus on the importance of a single word, or may only work when the number of words is limited and the  substitute space is small. 

%As Figure~\ref{fig:data} shows, for EHR data, the number of visit for each patient is not too large, and within each visit the number of ICD codes\footnote{\url{https://www.cdc.gov/nchs/icd/icd9cm.htm}} is also limited. These characteristics seem to fit the application of reinforcement learning-based models. However, the number of substitutes of a specific code may be huge according to the Clinical Classification Software,\footnote{\url{https://www.hcup-us.ahrq.gov/toolssoftware/ccs/AppendixASingleDX.txt}} which can lead to the failure of reinforcement learning-based models. On the contrary, score-based approaches can directly replace words based on the calculated saliency scores without dynamically searching the whole candidate space, which avoids the shortage of reinforcement learning-based approaches. Thus, it raises a question: \textit{Is it possible to ``hybridize'' those two kinds of methods for solving our problem?}

\smallskip
\noindent 
\textbf{Our Approach.} 
To solve these  challenges, in this paper we propose a new black-box adversarial attack method named~{\name} to explore the robustness of health risk prediction models. To cope with the unique EHR structure, {\name} adopts an approach that bridges score-based and reinforcement learning attacks, which has two steps including \textbf{hierarchical position selection} and  \textbf{substitute selection}. To be specific, the first step of~{\name} is to calculate the contribution score of \textit{each visit} by taking the temporal context into consideration. Next, it calculates the saliency score of \textit{each ICD code} within each visit. Using the contribution scores and saliency scores as the initialized policy parameters, it adopts a reinforcement learning framework to select the attacked positions in the EHR to generate globally optimized adversarial examples. After determining the attacked positions, {\name} selects the substitute  code that can bring the highest score change as the replacement of the original one.

\smallskip
\noindent 
\textbf{Contributions.}
To sum up, our contributions are as follows:
%\begin{itemize}%[leftmargin=*]
(1) To the best of our knowledge, we are the first to explore the robustness of health risk prediction models via black-box adversarial attacks. Compared with  white/gray-box setting, black-box adversarial attack is  more realistic, so our work can better approximate the robustness of real-world health risk prediction models.
(2) We propose a new black-box adversarial attack method called~{\name}, which is motivated to solve the challenges brought by the unique EHR data structure and alleviate the limitations of existing black-box adversarial attack techniques. It attacks the risk prediction models by taking the temporal context into consideration, and it can search better globally optimized adversarial examples by adopting a hybrid framework of reinforcement learning and score-based principles.
    % \item The experimental results show that {\name} can generate more adversarial examples to fool three representative risk prediction models on three real-world healthcare datasets compared with state-of-the-art black-box attack models, and it even performs better than white-box model LAVA~\cite{an2019longitudinal} on the Dementia dataset. In addition, we include a discussion on the defense against adversarial attacks for risk prediction models.
(3) We compare {\name} against state-of-the-art black-box adversarial attack methods in terms of attack success rate across  three real-world healthcare datasets. Results show that {\name} can generate more adversarial examples on average, and even outperforms white-box model LAVA~\cite{an2019longitudinal} in the Dementia dataset. With the experimental results, we further include a discussion on the defending against adversarial attacks for health risk prediction models. 

%% file: subfiles/relatedwork.tex
\section{Related Work}
%In this section, we will discuss the related work on adversarial attacks. Firstly, we cover the adversarial attack methods in the health domain and the role it plays in health risk prediction. %Next, among the existing adversarial attack work on different kinds of data~\cite{zhu2019robust, zugner2018adversarial,tang2020embarrassingly,chen2020rays,zang2020learning,ren2019generating,gao2018black, li2019textbugger, wang2020attackability},  
%We then discuss the black-box adversarial attack techniques that target at DNNs in the discrete text data. 

\subsection{Adversarial Attacks on EHR Data.}
The vulnerability problem is a vital issue for the  deep health risk prediction models for they are applied in the healthcare domain. Thus, we should understand how reliable health risk prediction DNNs are by exploring their robustness against adversarial attacks. However, adversarial attack research on EHR data is still in the early stage, and much more work is needed in designing effective adversarial attack methods in this domain, especially the black-box ones. To our knowledge, Sun et al.~\cite{sun2018identify} and An et al.~\cite{an2019longitudinal} are the ones who have explored the area of attacking DNNs with white/gray-box methods in the healthcare domain.~\cite{sun2018identify} proposes a white-box adversarial attack method for the EHR data that are described by continues values including vital signs and lab measurements, while~\cite{an2019longitudinal} conducts white-box and gray-box adversarial attacks on the ICD-based EHR data. Besides, Wang et al.~\cite{wang2020attackability} test an orthogonal matching pursuit-guided method for white-box evasion attack on the discrete EHR data. 

Nonetheless, the early work neglects the black-box adversarial attack setting, which is more realistic and challenging. Compared to white-box ones, black-box adversarial attack does not allow the attack methods to use the gradient information for adversarial example generation. Such a restriction makes the adversarial attack more difficult, and it is closer to the real-world scenario because the model gradient is unavailable in most cases. In this paper, we propose an effective black-box adversarial attack method which can improve the adversarial attack techniques for health risk prediction DNNs in a more realistic setting.% which can further improve model design and evaluation for this domain.

\subsection{Black-box Adversarial Attacks on Text.}
With the wide utilization of DNNs in various applications, people have become concerned about the reliability and vulnerability of DNNs in different fields. 
%The first work on exploring the vulnerability of DNNs can date back to~\cite{goodfellow2014explaining}, which discovers that DNNs can be fooled by adding small perturbation to the original input images. 
Among the existing adversarial attack work on different kinds of data including graphs~\cite{DBLP:conf/icml/DaiLTHWZS18} and images~\cite{goodfellow2014explaining, DBLP:conf/icml/IlyasEAL18,DBLP:conf/iclr/ChengLCZYH19,DBLP:conf/sp/ChenJW20,DBLP:conf/sp/Carlini017, DBLP:conf/iclr/KurakinGB17a}, text data~\cite{gao2018black,li2019textbugger,ren2019generating,zang2020learning,DBLP:conf/acl/EbrahimiRLD18} are the most relative ones to the EHR data because the search space of EHR and text data are both discrete. Thus, black-box text adversarial attack methods can be used as baselines in our experiments, including DeepWordBug~\cite{gao2018black}, TextBugger~\cite{li2019textbugger}, PWWS~\cite{ren2019generating} and a reinforcement learning method~\cite{zang2020learning}. Among them, DeepWordBug, TextBugger and PWWS can be categorized into score-based methods, which determine the attacked positions and perturbations by the saliency scores and substitute score changes, %A downside of score-based methods is that the position and substitute selection operations are deterministic, which may only generate local optimum adversarial examples.
and~\cite{zang2020learning} proposes a reinforcement learning technique that can dynamically try different kinds of substitute combinations. 
%it does not fit well with EHRs owing to their unique structure. 
As for our work,~{\name} aggregates the temporal context into a reinforcement learning framework to make it fit for EHR data, and it can be regarded as a hybrid method of the score-based and the reinforcement learning ones. Our experiments will show that such a hybrid design is a more effective adversarial attack solution for health risk prediction models in the black-box setting.

% In this paper, we will use the black-box adversarial attack methods that are originally proposed for the text classification models. This is because EHR can be regarded as a special document that can be understood after having the knowledge the encoding dictionary, where diagnosis codes are the ``words'' that constitutes the EHR ``documents''. The methods that we use as baselines include DeepwordBug~\cite{gao2018black}, TEXTBUGGER~\cite{li2019textbugger}, PWWS~\cite{ren2019generating} and a vanilla reinforcement learning (RL) method~\cite{zang2020learning}. DeepwordBug~\cite{gao2018black}, TEXTBUGGER~\cite{li2019textbugger} and PWWS~\cite{ren2019generating} can be categorized into score-based methods, which determines the attacked position and perturbation by the saliency scores and replacement change scores. A downside of score-based methods is that the position selection and replacement operations are deterministic, which may only find local optimum adversarial examples. To solve this problem,~\cite{zang2020learning} proposes a RL framework that can dynamically try different kinds of attack combinations. As for our work,~{\name} can be regarded as a hybrid method of the score-based one and the RL one. And our experiments will show that such a hybrid design is a more effective adversarial attack solution for risk prediction models.

%% file: subfiles/method.tex
\section{Methodology}
%In this section, we first define notations to formulate the adversarial attack task in the risk prediction application first, and then formally illustrate the proposed~{\name} framework. \cx{perhaps we dont need this paragraph}

\subsection{Problem Definition}

%\cx{I moved the original text into the defs/problems below. They need some polishing to be more formal.}

\begin{definition}[\textbf{Electronic Health Records}]
In our work, the EHRs of all patients are encoded by a high dimensional dictionary called ICD-9,~\footnote{\url{https://www.cdc.gov/nchs/icd/icd9.htm}} (International Classification of Diseases, Ninth Revision) and each symptom or abnormal finding is encoded into a unique code. In other words, each ICD-9 code is like a discrete symbol of the ICD-9 dictionary as an abstract of a unique medical symptom.
Mathematically, for a specific patient whose EHRs are denoted as $\mathbf{V}$, $\mathbf{V}$ is in the form of $[\mathbf{v}_1, \mathbf{v}_2, \cdots, \mathbf{v}_T]$, where $\mathbf{v}_t \ (1 \le t \le T)$ represents the result of visit $t$, and $T$ is the total number of visits. Each individual visit $\mathbf{v}_t = [c_1, c_2, \cdots, c_{n_t}]$ includes $n_t$ diagnosis codes encoded by the ICD-9 system. %Note that different from text data, the codes in $\mathbf{v}_t$ are unordered.
\end{definition}

\begin{problem}[\textbf{EHR Adversarial Attack}]
Let $F$ denote the health risk prediction DNN model. Given the input $\mathbf{V}$ of the patient and the corresponding ground truth label $y \in \mathcal{Y} = \{0, 1\}$, where $y = 1$ represents that patient will suffer from the target disease as a positive case and a negative one otherwise, in the training phase health risk prediction model $F$ is trained to generate a prediction score $\mu$ that is as close as to $y$, i.e.,  $\mu = F(\mathbf{V})$. Suppose that we have a test sample $\mathbf{V}_{\rm test}$ whose ground truth label is $y_{\rm test}$, and the classification threshold $\delta$. The output label given by $F$ will be 
$\hat{y} = {\rm sgn}(F(\mathbf{V}_{\rm test})-\delta)$, where ${\rm sgn}(x)=1$ if $x>0$ and ${\rm sgn}(x)=0$ otherwise. 

If the victim model $F$ can correctly predict the future status of the patient given the sample $\mathbf{V}_{\rm test}$, i.e., $\hat{y} = y_{\rm test}$, the target of adversarial attack is adding a perturbation $\Delta\mathbf{V}_{\rm test}$ to construct the adversarial example $  \mathbf{V}_{\rm test}^{\prime} = \mathbf{V}_{\rm test}+\Delta\mathbf{V}_{\rm test}$ such that $\mathbf{V}_{\rm test}^{\prime}$ can fool the victim model. That is, the perturbation makes the predicted label change as shown in Eq.~\eqref{eq:labelchange},
\begin{equation}
    \hat{y}^{\prime} = {\rm sgn}(F(\mathbf{V}_{\rm test}^{\prime})-\delta)  \ne  y_{\rm test},
    \label{eq:labelchange}
\end{equation}
where the perturbation $\Delta\mathbf{V}_{\rm test}$ should be as small as possible and is restricted by $||\Delta\mathbf{V}_{\rm test}|| < \epsilon$. We denote $||\Delta\mathbf{V}_{\rm test}||$ as the number of diagnosis code changes because EHR data is in a discrete space and $\epsilon$ as the maximum allowed attacks. We restrict the adversarial attack operation as substitution in our work because addition and deletion can be regarded substitutions related to a null character.
\end{problem}

\begin{figure}[t]
    \centering
    \includegraphics[width=0.5\textwidth]{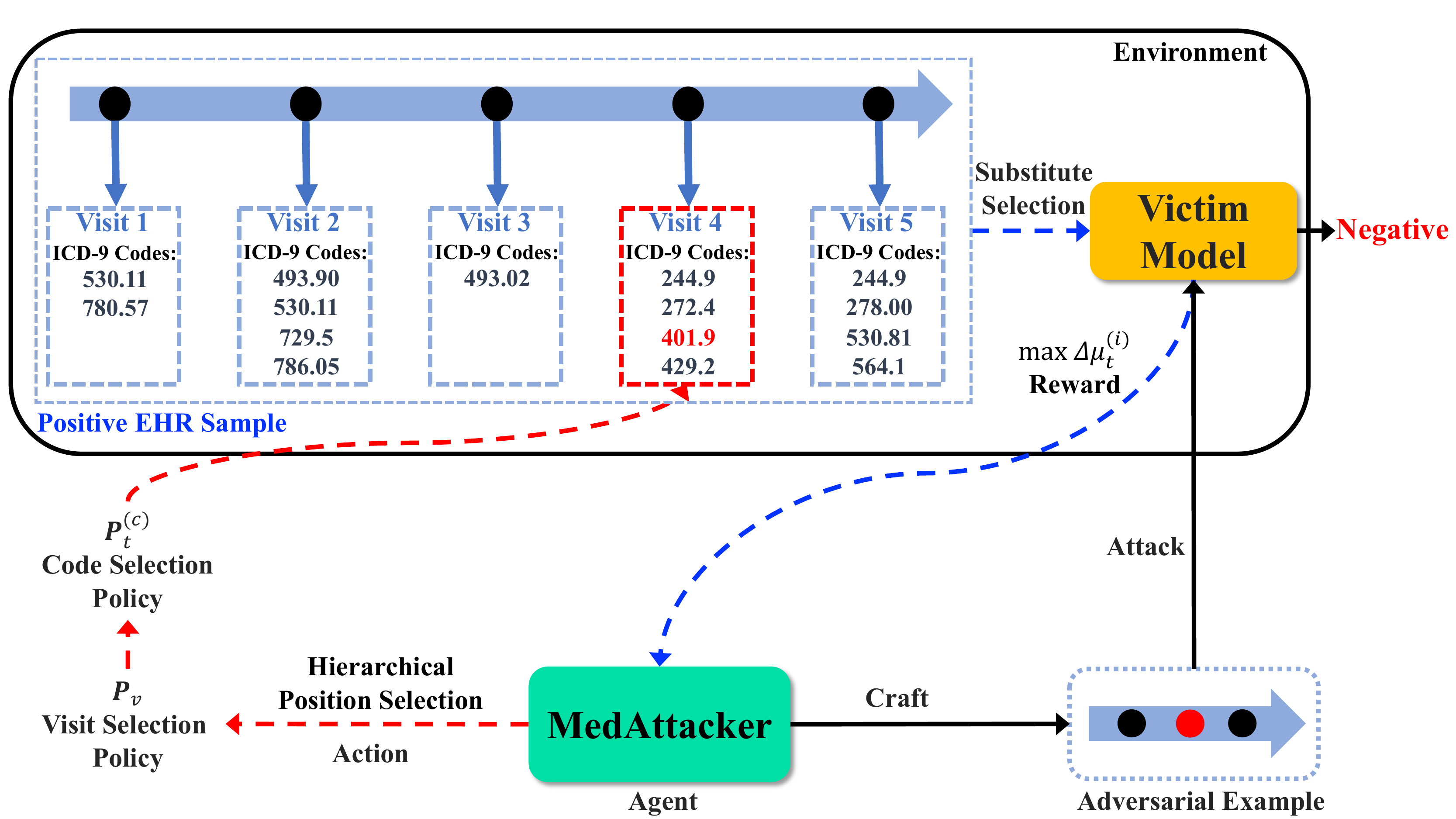} 
    \vspace{-0.1in}
    \caption{Overview of black-box adversarial attack on EHRs by~{\name}. {\name} is an agent who takes the action of position selection by the hierarchical position selection policy initialized with the temporal context, and it employs the substitute selection score changes output by the victim model as rewards to update the policy. Its learning task is to generate an adversarial example that can fool the victim model and cause it to change its prediction.}
    \label{fig:framework}
\end{figure}

\subsection{Proposed Method.}
%After discussing the procedures of position selection and substitute selection, we now show how they construct the proposed~{\name}. Since~{\name} selects the attacked position in a RL fashion, we would like to formulate it in the RL terms. Basically, the proposed~{\name} is the agent conducting EHR adversarial attack task. We can treat the EHR sample $\mathbf{V}$ as the environment and the initialized state $s_0 = \mathbf{V}$, and the actions $a$ are selecting the code to be attacked, whose policy is parameterized as by~{$\mathbf{\Theta}$}. One key problem is how we update the policy parameters in the process of generating adversarial sample, and our solution is use the maximum replacement score change $\max\Delta s_t^{(i)}$ in Eq.~\eqref{Eq: Update} as the reward $r$ to update the policy parameters $\mathbf{\Theta}$. 

%Following the previous discussion, there are two stages in EHR adversarial attack, i.e., position selection and substitute selection. As for~{\name}, its design is inspired by the existing black-box text adversarial attack methods. As mentioned earlier, existing state-of-the-art black-box text adversarial attack methods can be categorized into two types: score-based method and RL one. And~{\name} is a hybrid method which selecting attacked position by a RL framework and selecting substitute by changes of predicted scores. We will first introduce the details of the position selection and substitute selection processes in~{\name}, and then we show how to integrate them together in one framework.

%\textbf{Overview.} 
As shown in Figure~\ref{fig:framework}, the proposed~{\name} for adversarial EHR example generation in the black-box setting includes two steps, i.e., hierarchical position selection (selecting the positions of the attacked diagnosis codes by considering temporal context) and substitute selection (selecting the substitutes to replace the attacked diagnosis codes). %~{\name} complete these two steps in a hybrid fashion to lessen the limitations of existing black-box adversarial attack methods for EHRs. 
In the first step,~{\name} frames the position selection as a policy learned through reinforcement learning (RL). In this formulation,  the agent is {\name}, the environment consists of the EHR sample $\mathbf{V}$ and victim model $F$, and the state $s$ is represented by the EHR sample. Suppose it has $M$ learning episode to update the policy parameters, in each episode it will take several steps of actions. Due to the hierarchical characteristics of EHR data, i.e., \emph{code $\rightarrow$ visit $\rightarrow$ EHR}, {\name} will select the attacked visit firstly and the attacked diagnosis code within the visit later. They are then grouped as the action $a$ taken by the agent, and the policy is parameterized as $\mathbf{\Theta}$. To make use of the temporal context, we initialize the policy $\mathbf{\Theta}$ by the contribution scores of visits and saliency scores of diagnosis codes. As for the second step, we adopt a score-based fashion to determine the substitute code for the attacked position depending on which is the one that brings the maximum replacement score change in current state, which is integrated into the RL framework and harnessed as the reward $r$ for the agent~{\name} to update its policy. The details of these two processes are as follows.

\smallskip
\noindent\textbf{Hierarchical Position Selection.} 
The first step to generate an adversarial example is selecting the position of the attacked diagnosis codes. Existing black-box adversarial attack methods~\cite{ren2019generating,zang2020learning,gao2018black} on discrete data normally determine the attacked position directly by how much information will lose after removing the attacked word, but they neglect the context of the words within the same sentence, which leads to suboptimal results because existing health risk prediction models treat EHR data in a hierarchical way. %That is, health risk prediction models normally first extract the feature of each visit, and they then aggregate all visit features into a comprehensive feature for final prediction. 
Therefore, the position selection process in EHR adversarial attack should be conducted in a hierarchical way: \emph{selecting the attacked diagnosis code by firstly selecting the attacked visit and then deciding the attacked position within the selected visit.}

Thus, in our RL framework, this action is represented by two sets of parameters, and it is updated by the policy gradient~\cite{williams1992simple} framework. Without the loss of generality, suppose we have a positive test sample $\mathbf{V}$ as shown in Figure~\ref{fig:framework}, which can be correctly predicted by the trained model $F$, and $\mathbf{V} = [\mathbf{v}_1, \mathbf{v}_2, \cdots, \mathbf{v}_T]$ has information of $T$ visits. For the $i$-th visit $\mathbf{v}_t$, it has $n_t$ codes. Thus, the parameters to be learned include $\mathbf{p}_{\rm v} = [p_1^{({\rm v})}, ..., p_T^{ ({\rm v})} ]$, which is the probability distribution of selecting the \textbf{visit position}, and a group of parameters $\mathbf{p}_{\rm c} = \{\mathbf{p}_1^{ ({\rm c}) }, ..., \mathbf{p}_T^{ ({\rm c}) }\}$, which is the probability distribution of selecting the \textbf{code position} when the visit position is determined. For each $\mathbf{p}_t^{ ({\rm c}) }\in\mathbf{p}_{\rm c}$, $\mathbf{p}_t^{ ({\rm c}) } = [p_1^{(t)}, ..., p_{n_t}^{(t)} ]$, where $n_t$ is the number of codes in the $t$-th visit. Thus, the policy parameters to be learned in the policy gradient framework are $\mathbf{\Theta} = \{\mathbf{p}_{\rm v},  \mathbf{p}_{\rm c}\}$. In a learning episode,~{\name} selects the attacked visit $\mathbf{v}_t$ by sampling from $\mathbf{p}_{\rm v}$, and it then decides the attacked position by sampling from $\mathbf{p}_t^{({\rm c})}$.

The reason that we adopt a RL framework to select the attacked position is that it enables the adversarial example generation to be a stochastic process instead of a deterministic one, which can allow us to approximate the globally optimized adversarial example with more choices and give us more chances to successfully fool the trained model $F$. The difference between our framework and existing black-box RL framework~\cite{zang2020learning} is that ours will fit better with EHRs by selecting substitutes through replacement score change and taking the temporal context into consideration, which will be detailed as follows.

\smallskip
\noindent\textbf{Substitute Selection.} After we sample from the policy parameters $\mathbf{\Theta}$ and get the position that we are going to attack, the next step is to select a substitute to replace the attacked diagnosis code and generate the adversarial example. Suppose that the code to be attacked is $c_i$ from visit $\mathbf{v}_t$, and we denote $\mathbb{S}$ as the set of substitute codes.\footnote{We define set $\mathbb{S}$ as the set of codes in the same ICD-9 category of $c_i$ as the semantic constraints.} We then use the score changes brought by the substitutes to determine the substitute $c^{\prime}_i$ for $c_i$. That is, for each substitute code $c \in \mathbb{S}$, we can calculate the replacement score change by Eq.~\eqref{eq:scorechange},
\begin{equation}
    \Delta \mu_t^{(i)} = F((\mathbf{V}-c_i)\cup c) - F(\mathbf{V}),
    \label{eq:scorechange}
\end{equation}
where $((\mathbf{V}-c_i)\cup c)$ represents the EHR sample where $c_i$ is removed from $\mathbf{V}$ and replaced by $c$ in the attacked position. After obtaining all $\Delta \mu_t^{(i)}$ scores for every $c$ in set $\mathbb{S}$, we determine the best substitute code as demonstrated by Eq.~\eqref{Eq: substitute},
\begin{equation}
    c_i^{\prime} = \underset{c \in \mathbb{S}}{\arg\max}\ \Delta \mu_t^{(i)},
    \label{Eq: substitute}
\end{equation}
where $c_i^{\prime}$ is the code that we will finally employ to replace the attacked diagnosis code $c_i$.

In the model design, we determine the substitute code by Eq.~\eqref{Eq: substitute} instead of sampling by RL for it will be difficult to only use the $\max\Delta \mu_t^{(i)}$ to update position selection and substitute selection policy parameters simultaneously. We will have a detailed discussion on the policy update later.

%In our RL framework, this action is two sets of parameters, and it is updated by the policy gradient~\cite{williams1992simple} framework, a RL method, to determine the position to be attacked, which includes two actions, i.e., selecting a visit and selecting a code within the selected visit. Suppose we have a test sample $\mathbf{V}$ which can be correctly predicted by the trained model $F$, and $\mathbf{V} = [\mathbf{v}_1, \mathbf{v}_2, \cdots, \mathbf{v}_T]$. For the $i$-th visit $\mathbf{v}_t$, it has $n_t$ codes. Thus, the parameters to be learned include $\mathbf{p}_{\rm v} = [p_1^{({\rm v})}, ..., p_T^{ ({\rm v})} ]$, which is the probability distribution of selecting the visit position, and a group of parameters $\mathbf{p}_{\rm c} = \{\mathbf{p}_1^{ ({\rm c}) }, ..., \mathbf{p}_T^{ ({\rm c}) }\}$, which is the probability distribution of selecting the code position when the visit position is determined, and for each $\mathbf{p}_t^{ ({\rm c}) }\in\mathbf{p}_{\rm c}$, $\mathbf{p}_t^{ ({\rm c}) } = [p_1^{(t)}, ..., p_{n_t}^{(t)} ]$. Thus, the policy parameters to be learned in the policy gradient framework are $\mathbf{\Theta} = \{\mathbf{p}_{\rm v},  \mathbf{p}_{\rm c}\}$. 
\smallskip
\noindent\textbf{Policy Update.}
Given the framework above, the final design problem is how to update the  parameters $\mathbf{\Theta}$, which can be further divided into the sub-problems of how to initialize $\mathbf{\Theta}$ and how to define the rewards. 

\textbf{Parameter Initialization.}
Existing RL based black-box adversarial attack methods~\cite{zang2020learning} choose uniform distribution for initialization, but we find that using the~\textbf{contribution scores} and~\textbf{saliency scores} is a better way to initialize the policy $\mathbf{\Theta}$ for it takes temporal context into consideration. Specifically, we utilize the visit contribution scores to initialize $\mathbf{p}_{\rm v}$ and code saliency scores to initialize $\mathbf{p}_{\rm c}$. 

We define the visit contribution score as follows. To calculate the visit contribution of $\mathbf{v}_t$, we first calculate the output score given by the trained model $F$ when input is $[\mathbf{v}_1, ..., \mathbf{v}_t]$ and then calculate the output score when the input is $[\mathbf{v}_1, ..., \mathbf{v}_{t-1}]$. Next, we compute the contribution score $\xi_t$ by their difference,
\begin{equation}
    \xi_t = F([\mathbf{v}_1, ..., \mathbf{v}_t]) - F([\mathbf{v}_1, ..., \mathbf{v}_{t-1}]),
    \label{Eq: contribution}
\end{equation}
where $\xi_{t}$ indicate how much information that the whole visit of $\mathbf{v_t}$ can contribute to improving the health risk prediction given the context $[\mathbf{v}_1, ..., \mathbf{v}_{t-1}]$. By initializing $\mathbf{p}_{\rm v}$ as the normalized $[\xi_1, ..., \xi_T]$, the temporal context is utilized for determining the attacked position without hurting the stochastic property.

In terms of the saliency score of each code, it is calculated in a similar way as Eq.~\eqref{Eq: contribution}. For the $i$-th code $c_i$ in visit $\mathbf{v}_t$, we define the saliency score  as 
\begin{equation}
    \xi_t^{(i)} = F(\mathbf{V}) - F(\mathbf{V}-c_i),
    \label{Eq: saliency}
\end{equation}
where $(\mathbf{V}-c_i)$ denotes the incomplete EHR data where code $c_i$ is removed. The score $\xi_t^{(i)}$ is used to indicate the information that code $c_i$ possesses for health risk prediction. If score $\xi_t^{(i)}$ is high, it indicates that attacking $c_i$ can bring more salient influence. Thus, we initialize each $\mathbf{p}_t^{ ({\rm c}) }$ as the normalized $[\xi_t^{(1)}, ..., \xi_t^{(n_t)}]$, which fits with the unordered property of EHRs.

\textbf{Reward Calculation.}
We now discuss how to calculate reward in each learning episode.  Our solution is utilizing the maximum replacement score change $\max\Delta \mu_t^{(i)}$ in Eq.~\eqref{Eq: substitute} as the reward $r$ to update the policy parameters $\mathbf{\Theta}$, which enables us to integrate the position selection and substitute selection together and help~{\name} effectively find out the positions useful for adversarial example generation. Thus, in each learning episode, the total rewards of the adversarial example generation process is  $J(\mathbf{\Theta}) = \mathbb{E}(\sum_{\ell=0}^{\epsilon -1} \gamma^\ell r_\ell | \mathbf{\Theta}),$
%\begin{equation}
%    $J(\mathbf{\Theta}) = \mathbb{E}(\sum_{t=0}^{\epsilon -1} \gamma^t r_t | \mathbf{\Theta}),$
%    \label{Eq: Reward}
%\end{equation}
where $r_\ell$ is the reward attained in the step $\ell$, and $\gamma\in [0, 1]$ is the discount factor set to be 0.95. 

In addition, we update $\mathbf{\Theta}$ by the policy gradient method, in which the gradient of $J(\mathbf{\Theta})$ can be approximated by the REINFORCE algorithm~\cite{williams1992simple}, and the gradient $\nabla_{\mathbf{\Theta}} J(\mathbf{\Theta}) = \sum_{\ell=0}^{\epsilon -1} \nabla_{\mathbf{\Theta}} {\rm log}\pi_{\mathbf{\Theta}}(a_\ell|s_\ell)G_\ell,$ 
%\begin{equation}
%    \nabla_{\mathbf{\Theta}} J(\mathbf{\Theta}) = \sum_{t=0}^{\epsilon -1} \nabla_{\mathbf{\Theta}} {\rm log}\pi_{\mathbf{\Theta}}(a_t|s_t)G_t,
%    \label{Eq: Gradient}
%\end{equation}
where $a_\ell$ is the action that {\name} takes in step $\ell$, $s_\ell$ is the state of the environment, and $\pi_{\mathbf{\Theta}}(a_\ell|s_\ell)$ is the probability of taking action $a_\ell$ in state $s_\ell$, and $G_\ell$ refers to the discounted future reward $\sum_{\ell^{\prime}=\ell}^{\epsilon -1}\gamma^{\ell^{\prime}-\ell}r_{\ell^{\prime}}$. Given the approximated gradients, we can update the parameters according to $\mathbf{\Theta} \leftarrow \mathbf{\Theta} + \alpha \nabla_{\mathbf{\Theta}} J(\mathbf{\Theta})$,
%\begin{equation}
 %   , 
 %   \label{Eq: Update}
%\end{equation}
where $\alpha$ is the learning rate. 
%With the discussion above, we summarize  {\name}   as the pseudocode in Algorithm~\ref{Alg: MedAttack} .

%\begin{algorithm}

%\SetAlgoLined

 %\textbf{Inputs}: EHR sample to be attacked $\mathbf{V}$, victim model $F$, maximum allowed attacks $\epsilon$, and number of learning episodes $l$.
 %\begin{algorithmic}
 %\textbf{Output}: Adversarial example $\mathbf{V}^{\prime}$
 
 %Initialize policy parameters $\mathbf{\Theta}$ by Eq.~\eqref{Eq: contribution} and Eq.~\eqref{Eq: saliency}
 
% \While{{\rm episode} < l }{
  
 % Initialize the output candidate $\mathbf{V}^{\prime}$ as the original sample $\mathbf{V}$

  %\While{{\rm ||}$\Delta \mathbf{V}${\rm ||}< $\epsilon$}{
 % Choose the attacked visit $\mathbf{v}_t$ by sampling from $\mathbf{p}_{\rm v}$
  
  %Choose the attacked diagnosis code $c_i$ in $\mathbf{v}_t$ by sampling from $\mathbf{p}^{ ({\rm c}) }_t$ 
 
  %Select the substitute $c_i^{\prime}$ by Eq.~\eqref{Eq: substitute} 
  
  %Replace $c_i$ by $c_i^{\prime}$ in $\mathbf{V}^{\prime}$ and Update $\mathbf{V}^{\prime} \leftarrow (\mathbf{V}^{\prime} - c_i)\cup c_i^{\prime} $

  %\If{$F$ {\rm changes its prediction}}{
  % output the adversarial example $\mathbf{V}^{\prime}$
  % }
   
  %Update   $\mathbf{V} \leftarrow \mathbf{V}^{\prime}$
% }
 
  %Update the policy parameters $\mathbf{\Theta}$ by Eq.~\eqref{Eq: Reward},~\eqref{Eq: Gradient} and~\eqref{Eq: Update} }
 %\caption{{\name} for Black-Box EHR Adversarial Attack}
 %\label{Alg: MedAttack}
% \end{algorithmic}
%\end{algorithm}

%% file: subfiles/experiments.tex
\section{Experiments}

%In this section, we will show the experimental results of performance comparison between the proposed method and state-of-the-art baselines, model analysis by ablation study and scalability study, and a case study for illustrating adversarial example generation.

In this section we will show the experimental results for evaluating the effectiveness of the proposed method.% of performance comparison between the proposed method and state-of-the-art baselines, model analysis by ablation study and scalability study, and a case study for illustrating adversarial example generation.

\subsection{Experimental Setup}
\subsubsection{Datasets.}
In our experiments, we use three real-world health insurance claim datasets, including heart failure, kidney disease and dementia, which are collected by a health information technology company. %\footnote{The information is anonymized because of the double-blind review requirement.} 
The statistics of these datasets are shown in Table~\ref{Tab:Dataset}. %Note that in our experiments, we focus on the black-box adversarial attack problem, and the training data are only used to train each victim model. 
All the black-box adversarial attack experiments are conducted on the test data. The average number of visits on the heart failure, kidney disease, and dementia datasets is 38.74, 39.09, and 41.05, respectively. The average number of ICD codes per visit is  4.24, 4.40, and 4.71 on the heart failure, kidney disease, and dementia datasets, respectively. %In the test set, the minimum number of visits is 5 for each patient, and we vary the maximum number of accessible visits by attackers from 20 to 40.

\begin{table}[t]
\small
\caption{Statistics of the used datasets.}
\vspace{-0.1in}
\resizebox{0.47\textwidth}{!}{
\begin{tabular}{lcccc}
\toprule
Dataset &Heart Failure & Kidney Disease &Dementia\\ 
\midrule
Total Cases & 12,320 & 11,240 & 9,540\\
Positive Cases & 3,080 & 2,810 & 2,385\\
Negative Cases & 9,240 & 8,430 & 7,155\\
Test Set Size & 1,848 & 1,686 & 1,431\\
%Average Visits per Patient & 38.74 & 39.09 & 41.05\\
%Minimum Number of Visits & 5 & 5 & 5\\
%Maximum Number of Visits & 458 & 469 & 532\\
%Average Codes per Visit & 4.24 & 4.40 & 4.71\\
Unique ICD-9 Codes  & 8,692 & 8,802 & 7,813\\
\bottomrule
\end{tabular}
\label{Tab:Dataset}
}
\end{table}

% The statistics of the datasets we use that in the experiment are shown in Table~\ref{Tab:Dataset}. The target diseases of the datasets are heart failure, kidney disease and dementia, respectively, which is colected by a health information technology company\footnote{The information is anonymized because of the requirement of double-blind review.}. One thing we should note in Table~\ref{Tab:Dataset} is the size of the test set for it will be used to calculate the success rate after the attack, which is the metric we will use to quantify the effects of adversarial attack.

\subsubsection{Victim Models.}
Since we are conducting adversarial attacks on EHR data, we select three representative DNNs designed for health risk prediction task as the victim models in the adversarial attacks, which are Retain~\cite{choi2016retain}, SAnD~\cite{song2017attend} and HiTANet~\cite{luo2020hitanet}. The reason for selecting them is that Retain and SAnD are two state-of-the-arts that employ two mostly used temporal models in deep learning, i.e., recurrent neural networks (RNNs) and Transformer~\cite{vaswani2017attention}, and HiTANet is a method that emphasizes the utilization of time information, which is widely used~\cite{bai2018interpretable,baytas2017patient,luo2020hitanet} in risk prediction. %The following is their basic information.
%\begin{itemize}%[leftmargin=*]
%    \item \textbf{Retain}~\cite{chung2014empirical}%\footnote{\url{https://github.com/mp2893/retain}} 
%    \ is an early successful DNN that performs well in the health risk prediction task. It is built on a special kind of RNN called GRU~\cite{chung2014empirical}. It reverses the input EHR to model the way that physicians view the EHR in real world. It uses two different attention weights to detect important visits and clinical variables, which can also be used to interpret their importance.
 %   \item \textbf{SAnD}~\cite{vaswani2017attention}%\footnote{\url{https://github.com/khirotaka/SAnD}} 
 %   \ is the first health risk prediction model that uses Transformer structure for performance improvements. It uses the stacked Transformer to learn the long-term dependencies between all visit results, and it has a dense interpolation layer on the top to further aggregate the temporal information between different visits.
 %   \item \textbf{HiTANet}~\cite{luo2020hitanet}%\footnote{\url{https://github.com/HiTANet2020/HiTANet}} 
%    \ is the state-of-the-art health risk prediction model that is built upon the Transformer structure. One distinction between HiTANet and SAnD is that it emphasizes the utilization of time information associated with each visit, which allows it to better attach importance scores to different visits.
%\end{itemize}

\begin{table*}[ht]
%\small
\caption{Comparison on the number of successful attacks (the first row in each block) and success rate (the second row in each block). The average success rate is calculated for comparison when victim model is unknown.}

\resizebox{\textwidth}{!}{
\begin{tabular}{l|cccc|cccc|cccc}
\toprule

%Victim Model & \multicolumn{3}{c|}{HiTANet} & \multicolumn{3}{c|}{Retain} & \multicolumn{3}{c}{SAnD} \\ \hline
%Method    & Heart Failure     & Kidney       & Dementia   & Heart Failure       & Kidney        & Dementia & Heart Failure        & Kidney        & Dementia         \\ \hline

Dataset & \multicolumn{4}{c|}{Heart Failure} & \multicolumn{4}{c|}{Kidney Disease} & \multicolumn{4}{c}{Dementia} \\ \hline
Method    & HiTANet     & Retain       & SAnD  & Average & HiTANet     & Retain & SAnD & Average & HiTANet     & Retain       & SAnD  & Average   \\ \hline

\multirow{2}{*}{Random}    & 30          &     18        &   7   & 18.3  & 21     &      10    & 32  & 21.0  & 24 &  18 &    10  &  17.3    \\
& (1.62\%) & (0.97\%) & (0.38\%) & (0.99\%) &  (1.25\%) & (0.59\%) & (1.90\%) & (1.25\%) & (1.68\%) & (1.26\%) & (0.70\%)  & (1.21\%)
\\
\hline

\multirow{2}{*}{TextBugger}   & 216          &   119     &    4  & 113.0 &  182    &    138        &   104  & 141.3 & 117    & 109  & 6  & 77.3 \\

& (11.69\%) & (6.44\%)& (0.22\%) & (6.11\%) & (10.79\%)& (8.19\%)& (6.17\%) & (8.38\%) & (8.18\%)& (7.62\%)& (0.42\%) & (5.40\%)
\\
\hline

%\multirow{2}{*}{DeepWordBug}  & 231          &  248           &    147   & 113        &  96          &   87   & 12    & 92   & 15  \\
%& (12.50\%) & (14.71\%) & (10.27\%) & (6.11\%) & (5.69\%) & (6.08\%) & (0.65\%) & (5.46\%) & (1.05\%)

\multirow{2}{*}{DeepWordBug}  & 231         & 113           & 12  & 118.7  &  248        &  96         & 92 & 145.3  &   147   &   87    & 15 & 83.0\\
& (12.50\%)& (6.11\%) & (0.65\%) &(6.42\%) & (14.71\%) & (5.69\%)& (5.46\%)  &(8.62\%) & (10.27\%) & (6.08\%) & (1.05\%) & (5.80\%)

\\ 
\hline

\multirow{2}{*}{PWWS-Saliency}     & 277          &    129        &  48  & 151.3 & 264        &  98   & 229  &  197.0 & 277  & 166  &  66    & 169.7  \\
& (14.99\%) & (6.98\%) & (2.60\%) & (8.19\%)  & (15.66\%) & (5.81\%) & (13.58\%)& (11.68\%)&  (19.36\%) &(11.60\%) & (4.61\%) & (11.86\%)
\\
\hline
%\multirow{2}{*}{PWWS} & 369          &   332         &   359   & 162         &   \textbf{154} & 204 & 52 & \textbf{239} & \textbf{77}     \\
%& (19.97\%) & (19.69\%) & (25.09\%) & (8.77\%) & \textbf{(9.13\%)} & (14.26\%) & (2.81\%) & \textbf{(14.18\%)} & \textbf{(5.38\%)}
%\\

\multirow{2}{*}{PWWS} & 369          & 162            & 52  & 194.3  &   332          &   \textbf{154} & \textbf{239} &  241.7  &  359 & 204 & \textbf{77}  &  213.3  \\
& (19.97\%) & (8.77\%)& (2.81\%) & (10.52\%) & (19.69\%) & \textbf{(9.13\%)} & \textbf{(14.18\%)} & (14.33\%)  & (25.09\%) & (14.26\%)& \textbf{(5.38\%)} & (14.91\%) 
\\
\hline
%\multirow{2}{*}{RL-Attack} & 347         &      301     &  272     & 146      &   132         &  160   & 25 &    142 & 30      \\
%& (18.78\%) & (17.85\%) & (19.01\%) & (7.90\%) & (7.83\%) & (11.18\%) & (1.35\%)& (8.42\%) & (2.10\%)
%\\

\multirow{2}{*}{RL-Attack} & 347        & 146     & 25  &  172.7 &      301     &   132         &    142  &  191.7 &  272  &  160  & 30    &  154.0 \\
& (18.78\%) & (7.90\%) & (1.35\%)& (9.34\%) & (17.85\%) & (7.83\%) & (8.42\%) & (11.37\%)& (19.01\%)& (11.18\%) & (2.10\%) & (10.76\%)
\\

\hline 
%\multirow{2}{*}{{\name}}
%&     \textbf{426}     &      \textbf{384}       &    \textbf{369}   &  \textbf{166}     & 149           &  \textbf{210} & \textbf{64}  &  218 & 63             \\ 
%&   (\textbf{23.05\%}) & (\textbf{22.78\%}) & (\textbf{25.79\%}) & (\textbf{8.98\%}) & (8.84\%) & (\textbf{14.68\%}) & (\textbf{3.46\%}) & (12.93\%) & (4.40\%)\\

\multirow{2}{*}{{\name}}
&     \textbf{426}     &  \textbf{166}     & \textbf{64}  & \textbf{218.7}  &      \textbf{369}      & 149            &  218 &  \textbf{245.3}  &    \textbf{384}  &  \textbf{210} & 63    &   \textbf{219.0}     \\ 
&   (\textbf{23.05\%})  & (\textbf{8.98\%}) & (\textbf{3.46\%}) & (\textbf{11.83\%}) & (\textbf{21.89\%}) & (8.84\%) & (12.93\%) & (\textbf{14.55\%})  & (\textbf{26.83\%})  & (\textbf{14.68\%}) & (4.40\%) & (\textbf{15.30\%}) \\

%\cline{2-10}
%&   \multicolumn{3}{c|}{Avg: \textbf{11.83\%}} & \multicolumn{3}{c|}{Avg: \textbf{14.85\%}} & \multicolumn{3}{c}{Avg: \textbf{14.96\%}}\\

\bottomrule

\end{tabular}
\label{Tab: Comparison}

}
\end{table*}

\subsubsection{Baselines.}
%After describing the target models that we attempt to attack, we now discuss the black-box adversarial attack baselines that we use in the experiments. A naive adversarial attack baseline is randomly selecting the attacked positions and substitutes. And we term it as~\textbf{Random}. There are four state-of-the-art black-box adversarial attack baselines that craft the adversarial samples in a deterministic process, which are DeepWordBug~\cite{gao2018black}, TextBugger~\cite{li2019textbugger}, PWWS~\cite{ren2019generating} and PWWS-Saliency. We also include a recent reinforcement learning (RL) method~\cite{zang2020learning} that generate adversarial samples in a stochastic way, and we term it to \textbf{RL-Attack}. Here is their basic description.
Since we are the first to work on the adversarial attack on EHRs, there is few baselines specifically designed for the EHR adversarial attack. Therefore, most baselines that are used in the experiments are originally designed for the text adversarial attack. To make the input data fit the text adversarial attack models, we treat each visit as a sentence and treat each code as a word. %Because diagnosis codes within each visit are unordered in the EHR, we treat the appearing order of the codes in EHR as the order of words in a sentence for the baselines. Note that for the fair comparison, the order of codes within each visit is fixed among all approaches.
In our experiments, we use six baselines, including a naive approach and five (including four score-based and one reinforcement learning-based) state-of-the-art \textbf{black-box adversarial attack methods} as follows:(1) \textbf{Random},  a naive adversarial attack baseline which randomly selects the attacked positions and substitutes; (2) \textbf{DeepWordBug}~\cite{gao2018black}; (3) \textbf{TextBugger}~\cite{li2019textbugger};  (4) \textbf{PWWS}~\cite{ren2019generating} and its  varient (5) \textbf{PWWS-Saliency} which only uses the saliency score to determine which word to be attacked; and (6) \textbf{RL-Attack}~\cite{zang2020learning}.
In addition, we use the state-of-the-art \textbf{white-box} EHR adversarial attack method \textbf{LAVA}~\cite{an2019longitudinal} which can access the model gradients as a baseline to better understand the attack effect of black-box adversarial attacks. %This will be helpful for better understanding black-box adversarial attacks.

\subsubsection{Implementation.} 
Our experiments are implemented in the PyTorch framework in the hardware environment of a NVIDIA Tesla P100 GPU and Intel Xeon E5-2680 CPUs. The reinforcement learning environment is implemented in the OpenAI Gym~\cite{brockman2016openai} package, and the learning rate of the policy parameters is $1\times 10^{-3}$. When implementing our algorithm, we set the hyperparameter $l=500$. The set of $\mathbb{S}$ for each code is made up of diagnosis codes in the same ICD category by the Clinical Classification Software-DIAGNOSES. %\footnote{\url{https://www.hcup-us.ahrq.gov/toolssoftware/ccs/AppendixASingleDX.txt}}
We set size of $\mathbb{S}$ no more than 10 for efficiency reason and for each category they are selected randomly. As for baselines, the substitute selection of the score-based methods are set as the same way that PWWS does for fair comparison, which is decided by the $\Delta P$ score. %And we implement LAVA by its official codes.
We vary the the maximum allowed attacks $\epsilon$ from 5 to 15.

\subsubsection{Evaluation Metrics.}
The main goal of attacking victim models is to change the labels of test set data that are correctly predicted by the black-box DNNs, so 
the first metric is \textbf{the number of successful attacks} that each method make, and dividing it by the size of test set can get the \textbf{success rate}~\cite{zang2020learning,li2019textbugger}, which shows the covering range of successful attacks and the decreased accuracy of the victim model. We can then calculate the \textbf{average success rate} for one dataset across different health risk prediction models to estimate the attack result when the attacked model is unknown as a black-box. The larger they are, the better the performance is.

\subsection{Performance Evaluation.}

Since there is always physical restriction on the access of EHR data in real-world defense, we first validate the performance of models under a relatively restrict and realistic setting : (1) The maximum accessible visits allowed for attack in the test set is 20, which is about a half of average numbers of visits of three datasets, and (2) the maximum allowed attacks $\epsilon =5$. Such a setting can better validate the effectiveness of designed method for testing the robustness of health risk prediction models.
The comparison of adversarial attack results between~{\name} and baselines are shown in Table~\ref{Tab: Comparison}. From Table~\ref{Tab: Comparison}, we can see that~{\name} can always attain the highest average success rate across three test datasets, which is 11.83\%, 14.55\% and 15.30\% in the heart failure, kidney disease and dementia dataset, respectively. Specifically, it achieves the best performance over three datasets against three different victim models in 6 out of 9 cases. These results demonstrate that~{\name} has the best generalization ability for EHR adversarial attack compared to existing black-box adversarial attack techniques, and such a versatile attack ability indicates that~{\name} can be well applied in the black-box setting \emph{when victim models are unknown}.

Besides,~{\name} is especially good at dealing with difficult situations when attacking in relatively large datasets or against victim models using time information. Among all the three datasets,~{\name} always has the highest success rate in the largest data, i.e., heart failure, against different victim models. Compared to the other two datasets, the test samples in heart failure are more varied and diverse. 
Thus, in this complex situation,~{\name} is able to bring out the potential of reinforcement learning and generate more adversarial examples to successfully fool the victim models.  Moreover,~{\name} can constantly achieve the best attack success rate when the victim model is HiTANet, which employs time information for health risk prediction, and the success rate is 3.08\%, 2.20\%, and 1.74\% higher than the second best method PWWS in heart failure, kidney disease and dementia dataset, respectively. This is mainly because the hierarchical position selection strategy is useful for~{\name} to attack health risk prediction models using time information by discovering most contributory visits for time embedding learning, which is an advanced design~\cite{bai2018interpretable,baytas2017patient,luo2020hitanet} widely used in health risk prediction.

\begin{table}[t]
\caption{Comparison with LAVA on success rate.}
\begin{tabular}{l|ccc}
\toprule
Victim Model & \multicolumn{3}{c}{Retain}  \\ \hline
Method    & Heart Failure     & Kidney       & Dementia            \\ \hline
%\multirow{2}{*}{LAVA} & \textbf{228}         &  \textbf{183}     &  163           \\ & \textbf{12.34\%} & \textbf{10.85\%} & 11.39\% \\
LAVA & \textbf{12.34\%} & \textbf{10.85\%} & 11.39\% \\
\hline
%\multirow{2}{*}{{\name}}    &     166     & 149           &  \textbf{210} \\ & 8.98\% & 8.84\% & \textbf{14.68\%} \\
{\name}    &     8.98\% & 8.84\% & \textbf{14.68\%} \\
\bottomrule
\end{tabular}
\label{Tab: Comparison with LAVA}
\end{table}

\textbf{Discussion.} The results above show that~{\name} can alleviate the limitations of reinforcement learning and score-based principles to have better generalization ability while conducting attacks in complex situations. Compared to score-based methods~\cite{gao2018black,li2019textbugger,ren2019generating}, ~{\name} selects attacked code positions in a stochastic fashion by reinforcement learning, which can help~{\name} generate more effective adversarial examples to fool the victim models. In addition, compared to RL-Attack~\cite{zang2020learning},~{\name} initializes its parameters to take the temporal context into consideration and is only required to learn the position selection parameters, which can help it find out better substitutes for adversarial attack. Hence,~{\name} is more suitable for EHR adversarial attacks.

\subsection{Comparison with White-box Attack.}
To evaluate the effectiveness of black-box adversarial attacks on EHRs, we  compare the attack results between black-box adversarial attacks and white-box ones. Thus, we also employ LAVA as a baseline for white-box adversarial attack, and the experimental results are listed in Table~\ref{Tab: Comparison with LAVA}.  
We compare them in the case of Retain owing to the availability of official implementation codes of LAVA. Because white-box ones have the knowledge of gradients, it is not surprising to see that LAVA has better performance on the datasets of heart failure and kidney disease. But compared to LAVA, {\name} can still have 72.77\% and 81.47\% adversarial attack effects of the white-box method on heart failure and kidney disease datasets, respectively, which demonstrates the effectiveness of the designed method.

Most importantly, {\name} can have better attack results on the dementia dataset. This indicates that for a relatively small dataset such as the dementia one, black-box adversarial attack techniques have the potential to achieve better results against white-box ones because the size of the dataset may restrict the model parameter learning, which may mislead the white-box attack result owing to inaccurate gradients.

% \begin{table}[t]
% \small
% \caption{Model analysis on module design.}
% \begin{tabular}{c|ccc}
% \toprule

% Target Model & \multicolumn{3}{c}{HiTANet}  \\ \hline
% \multirow{2}{*}{Modules} & w/o   Substitute    &  w/o  Score-based    &  \multirow{2}{*}{{\name}}           \\ &  Selection  & Initialization  \\
% \hline
% Success Rate   &     18.02\%     & 22.19\%           &  \textbf{23.05\%} \\ 
% %\multirow{2}{*}{Success Rate}   &     18.02\%     & 22.19\%           &  23.05 \\ 
% %& 8.98\% & 8.84\% & \textbf{14.68\%} \\
% \bottomrule

% \end{tabular}
% \label{Tab:module}
% \end{table}

\begin{table}[t]
\small
\caption{Model analysis on module design.}
\begin{tabular}{l|cc}
\toprule

%Target Model & \multicolumn{3}{c}{HiTANet}  \\ \hline
Victim Model & HiTANet  \\ \hline
Modules & Success Rate \\ \hline
w/o Substitute Selection & 18.02\%\\
w/o Hierarchical Position Selection & 20.83\%\\
w/o Score-based Initialization & 22.19\% \\

{\name} & \textbf{23.05\%} \\ 

% \multirow{2}{*}{Modules} & w/o   Substitute    &  w/o  Score-based    &  \multirow{2}{*}{{\name}}           \\ &  Selection  & Initialization  \\
% \hline
% Accuracy Decrease   &     18.02\%     & 22.19\%           &  \textbf{23.05\%} \\ 
%\multirow{2}{*}{Accuracy Decrease}   &     18.02\%     & 22.19\%           &  23.05 \\ 
%& 8.98\% & 8.84\% & \textbf{14.68\%} \\
\bottomrule

\end{tabular}
\label{Tab:module}
\end{table}

\subsection{Model Analysis. \label{Sec: model_analaysis}}
Without the loss of generality, we conduct three groups of model analysis experiments on the heart failure disease against HiTANet.%including the ablation study on module design and scalability studies on maximum allowed attacks and maximum accessible visits. .
%, and the results are shown in Table~\ref{Tab:module} and Table~\ref{Tab:attack and visit}.  

\subsubsection{Ablation Study.}
%\textcolor{blue}{rewrite this paragraph}
To validate the module design, as shown in Table~\ref{Tab:module} we utilize three variants of {\name} for comparison  in ablation study experiment. From Table~\ref{Tab:module} we can find that if {\name} selects substitutes stochastically (``w/o Substitute Selection'') as RL-Attack  does, the performance will decrease by 5.03\%, which shows that we can release its limitation by selecting substitute in a score-based fashion as Eq.~\eqref{Eq: substitute} indicates. As for the position selection strategy, if {\name} does not select the attacked position in a hierarchical way ( ``w/o Hierarchical Position Selection''), the attack effect on success rate will decrease by 2.22\%. Hence, the hierarchical position selection is more suitable for the EHR structure. In addition, if {\name} initializes its policy uniformly (``w/o Score-based Initialization''), its performance will decrease from 23.05\% to 22.19\%. Thus, we should initialize the policy parameters by the contribution and saliency scores for it aggregates position selection with the temporal context.

%\begin{table}[t]
%\footnotesize
%\caption{Model Analysis on Maximum Allowed Attacks and Maximum Accessible Visits}
%\begin{tabular}{c|ccc|ccc}
%\toprule
%Target Model & \multicolumn{6}{c}{HiTANet}  \\ \hline
%Maximum   &\multirow{2}{*}{5}    & \multirow{2}{*}{10}    &  \multirow{2}{*}{15}  & \multirow{2}{*}{5} & \multirow{2}{*}{5}   &\multirow{2}{*}{5}      \\ Allowed  Attacks & \  & \ & \ & \ & \ \\
%\hline
%Maximum  &\multirow{2}{*}{20}    & \multirow{2}{*}{20}    &  \multirow{2}{*}{20} &\multirow{2}{*}{20}   & \multirow{2}{*}{30}    &  \multirow{2}{*}{40}        \\ Accessible Visits & \  & \ & \ & \ & \ \\
%\hline

%Accuracy Decrease   &     23.05\%     & 27.92\%           &  \textbf{30.52\%}  &      23.05\% & 28.35\% & \textbf{31.33\%}\\ 

%\bottomrule

%\end{tabular}
%\label{Tab:attack and visit}
%\end{table}

\begin{figure}[!t]
\centering
\subfigure{
\begin{minipage}[b]{0.23\textwidth}
  \includegraphics[width=1\textwidth]{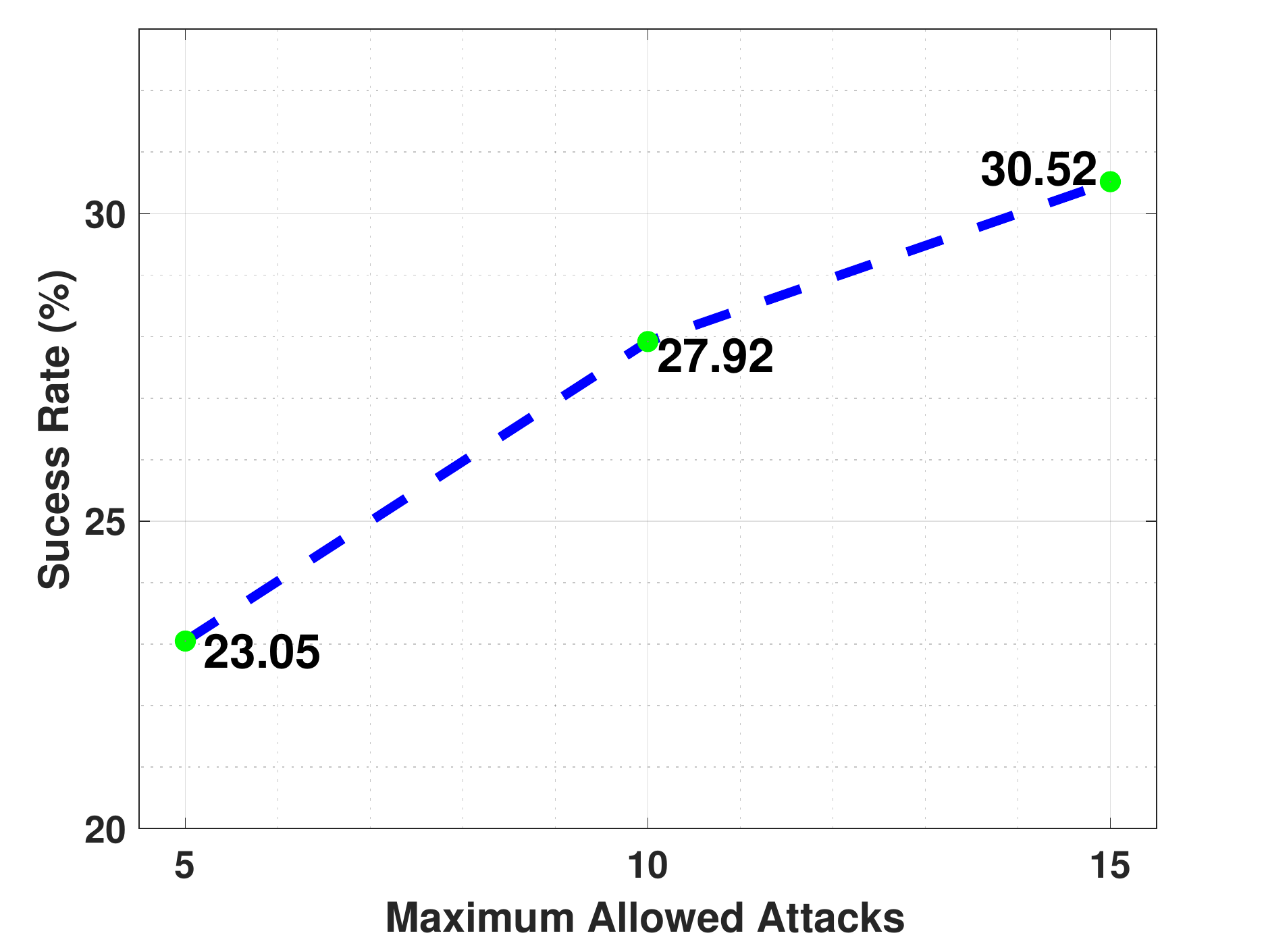} 
  \end{minipage}}
\subfigure{
\begin{minipage}[b]{0.23\textwidth}
  \includegraphics[width=1\textwidth]{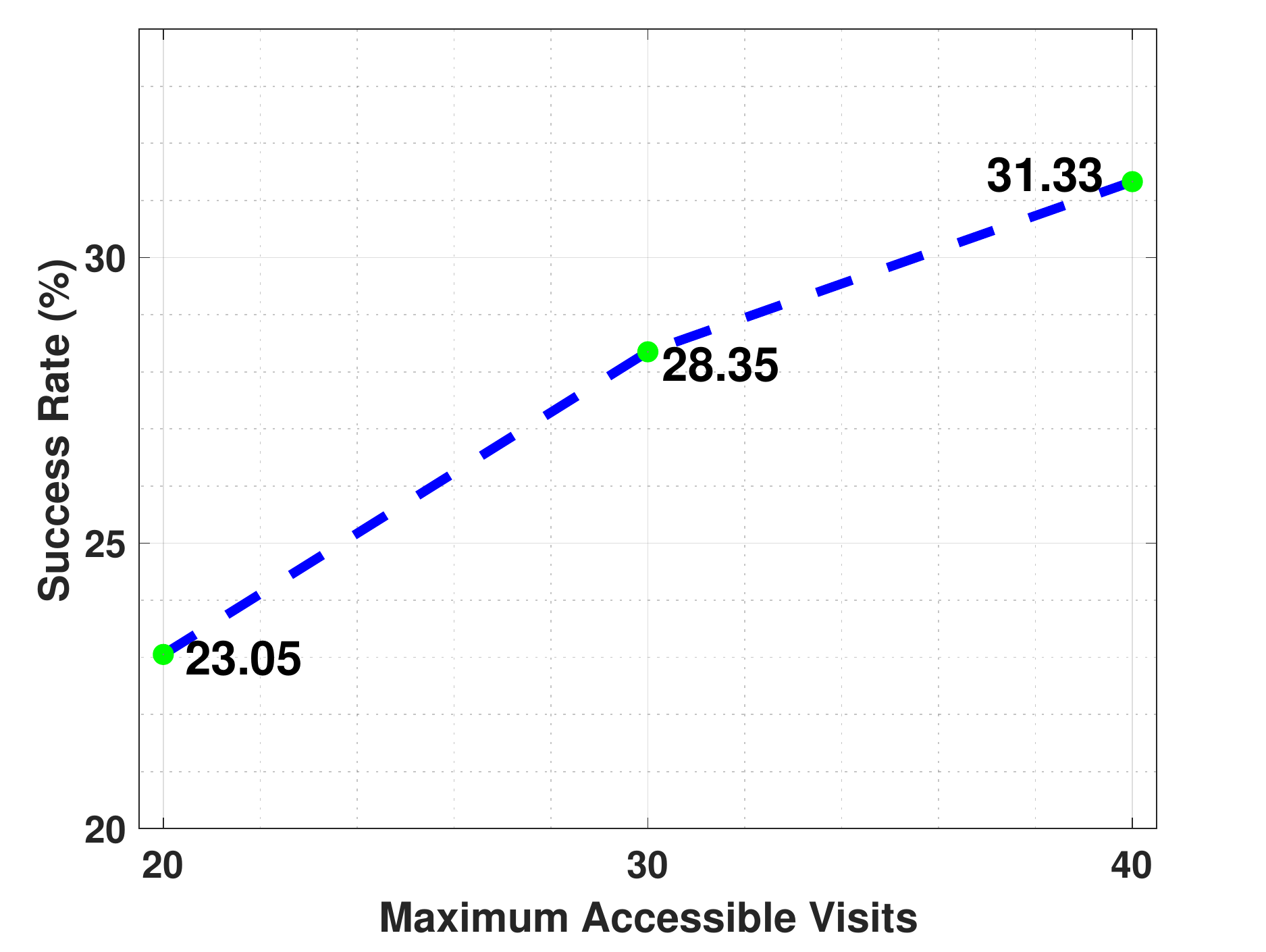}
  \end{minipage}}
  %\vspace{-0.2in}
\caption{(a) Scalability study on maximum allowed attacks. (b) Scalability study on maximum accessible visits. } %\guanjie{will the acc decrease keep rising with more allowed attacks? what are the performance of other methods under more attacks? if they can perform similarly as us under more attacks, why do we choose to limit the attack number at 5? same as the maximum accessible visits.}
    \label{fig:attack and visit}
\end{figure}

\subsubsection{Scalability Study.}
%We also conduct scalability study on maximum allowed attacks because the number of successful attacks should be positively correlated with the number of maximum allowed attacks for a ideal adversarial attack design. 
To validate the model scalability, we conduct an experiment by changing the maximum allowed attacks from 5, 10 to 15, while keeping the maximum accessible visits as 20. The experimental results are shown in Figure~\ref{fig:attack and visit}(a). From the results we can see that if we allow {\name} to conduct more attacks on the victim model, the accuracy will keep increasing from 23.05\% to 30.52\%, which satisfies the scalability requirement that the performance of {\name} can keep improving if it is allowed to conduct more adversarial attacks.

%In addition, a good design should also improve its success rate if more EHR data of the same patient is exposed. 
After that, we keep the the maximum allowed attacks as 5 and try to see the influence of maximum accessible visits by increasing it from 20 to 30 and to 40, and the results are shown in Figure~\ref{fig:attack and visit}(b). We can observe a rising trend of the attack result as the maximum accessible visits increase, which increases to 31.33\% when {\name} can access up to 40 visits of each patient. The reason is that the larger is the number of maximum accessible visits, the easier it is for {\name} to find the sensitive positions.
Thus, {\name} can make more successful attack attempts if more visits within the same EHR sample is available.

\subsection{Case Study.}

\begin{figure}[t]
    \centering
    %\hspace*{-1cm}
    \includegraphics[width=0.5\textwidth]{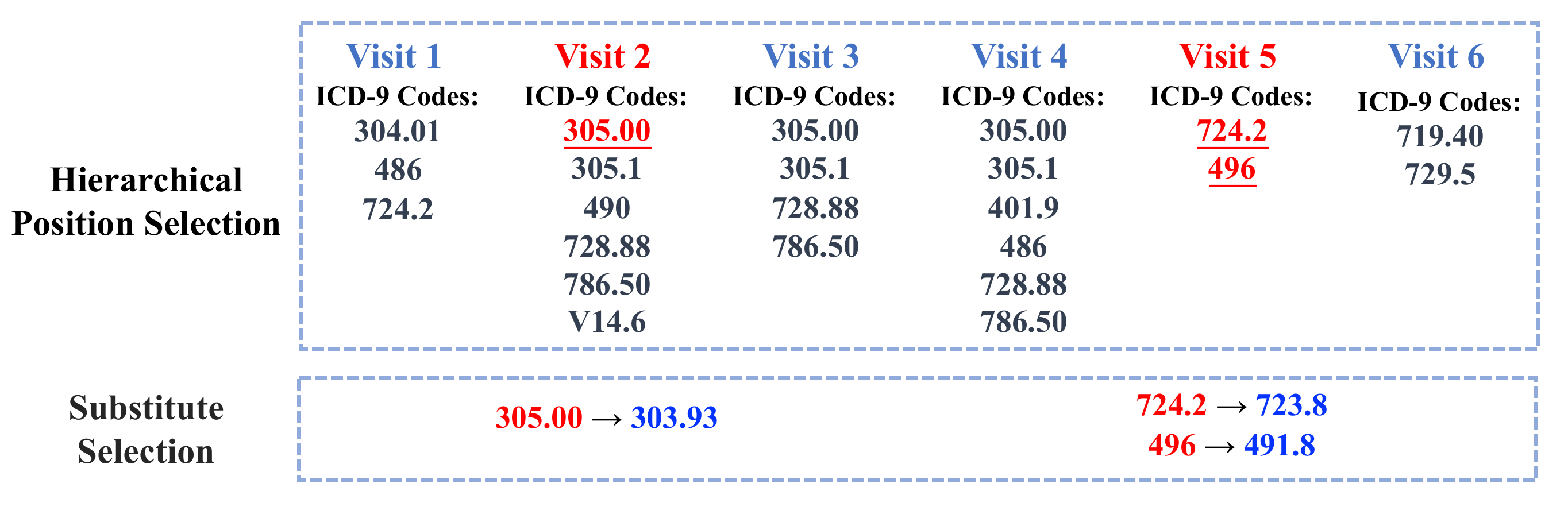} 
    %\vspace{-0.1in}
    \caption{Illustration of case study. The selected attacked positions are  colored in red, and the selected substitutes are colored in blue.}
    \label{case_study}
\end{figure}

We also include a case study as follows to further illustrate the adversarial attack process conducted by~{\name}. As Figure~\ref{case_study} shows, given a positive EHR sample in the heart failure dataset whose predicted score of being a positive case by HiTANet is 0.70,~{\name} selects code ``305.00'' (\emph{alcohol abuse}) from Visit 2, code ``724.2'' (\emph{lumbago}) and code ``496'' (\emph{chronic airway obstruction}) from Visit 5 to construct the adversarial example. Note that these symptoms are closely related to the heart failure disease, %\footnote{\url{https://www.mayoclinic.org/diseases-conditions/heart-failure/symptoms-causes/syc-20373142}}  
so the hierarchical position selection process is helpful in finding the diagnosis codes that are closely relevant with the target heart failure disease. Furthermore, since the substitute set of each code is restricted within the same category in the ICD-9 coding system, the semantic in the adversarial example is similar to the original one. For instance, substitute ``723.8'' (\emph{cervical syndrome}) and the original code ``724.2'' are both in the category of ``Spondylosis'', which can ensure the adversarial example still looks reasonable by humans and can better test the robustness of victim models. After the attack, HiTANet downgrades the predicted score to 0.33 and predicts it as a negative case, which shows the effectiveness of ~{\name} in fooling deep health risk prediction models.

\subsection{Discussion on Health Risk Prediction Defense.}

After finding an effective black-box adversarial attack method, based on the experimental results above, we provide the following suggestions on defending against EHR adversarial attack in real-world applications: for one thing, \emph{restricting the visibility of EHR data.} Since only allowing 5 attacks on 20 visits of the EHR can cause the victim model to decrease its accuracy up to 26.83\% by the proposed method, we should design effective mechanism for protecting EHR data. %From Figure~\ref{fig:attack and visit}, we can see if we allow {\name} to access more EHR data and conduct more attacks, the damage to the health risk prediction performance will be worse. Thus, the first thing we can do to protect health risk prediction models is restricting the visibility of EHR data. 
For another, \emph{improving health risk prediction model design.} %Table~\ref{Tab: Comparison} shows that for the same dataset, the performance of successful attack by {\name} can be different depending on the victim model. 
We recommend using the Transformer structure as building blocks for the attack success rates caused by {\name} in the heart failure and dementia dataset are both less than 5\% against SAnD but they are 8.98\% and 14.68\% against Retain, respectively. This is because Transformer processes the temporal data in parallel, which can reduce the error accumulation~\cite{holden2017phase} problem in RNNs and improve model robustness.% and make it more robust. %We also want to mention that health risk prediction model should be careful about utilizing the time information. Although HiTANet and SAnD both harness Transformer as the building block, the performance decrease is more severe when HiTANet is used as the victim model. This is probably because HiTANet  requires more on time information for health risk prediction, and it is more likely for a small perturbation to prevent it from learning good visit embeddings by affecting the time information feature. Thus, we suggest time information should be processed separately from visit information to improve health risk prediction model design.

%% file: subfiles/conclusions.tex
\section{Conclusion}
The robustness of health risk prediction models is crucial because they are related to human lives. Although researchers have investigated their vulnerability by the white/gray-box adversarial attacks, a more realistic stimulation of real-world adversarial attacks, i.e.,  EHR adversarial attack in the black-box setting, has not been explored yet. To increase the momentum in this field, we introduce a black-box adversarial attack framework named~{\name} to explore the robustness of health risk prediction models. %which can alleviate the limitations of the score-based and reinforcement learning methods in black-box adversarial attacks. 
It is more suitable for adversarial attack on EHR data in the black-box setting because it takes the temporal context of EHR into consideration, and the stochastic position selection and deterministic substitute selection processes can help it better generate globally optimized adversarial examples.~{\name} can achieve the highest average success rate in three real-world datasets against three representative health risk prediction models compared with the state-of-the-art baselines. Besides,  in  a relatively small EHR dataset, it can even outperform the white-box EHR adversarial attack baseline. Finally, based on our results we include a discussion to help improve the defense mechanism of health risk prediction models.